\documentclass[10pt,twocolumn,letterpaper]{article}

\usepackage{iccv}
\usepackage{times}
\usepackage{epsfig}
\usepackage{graphicx}
\usepackage{amsmath}
\usepackage{amssymb}

\usepackage[breaklinks=true,bookmarks=false]{hyperref}

\iccvfinalcopy 

\usepackage{hhline}
\usepackage{colortbl}
\usepackage{multirow}
\usepackage{array}
\definecolor{col1}{RGB}{243, 230, 195}
\definecolor{col2}{RGB}{208, 220, 242}
\definecolor{Gray}{gray}{0.9}
\definecolor{firstcolor}{RGB}{252, 213, 206}
\definecolor{secondcolor}{RGB}{252, 246, 189}

\begin{document}
\title{CVRecon: Rethinking 3D Geometric Feature Learning For Neural Reconstruction}

\author{\large \href{https://ziyue.cool}{Ziyue Feng}\textsuperscript{1} \quad
Liang Yang\textsuperscript{2} \quad
Pengsheng Guo\textsuperscript{3} \quad
Bing Li\textsuperscript{1} \\
 \textsuperscript{1}Clemson University \quad \textsuperscript{2}City University of New York \quad \textsuperscript{3}Carnegie Mellon University
}

\maketitle

\begin{abstract}
   Recent advances in neural reconstruction using posed image sequences have made remarkable progress. However, due to the lack of depth information, existing volumetric-based techniques simply duplicate 2D image features of the object surface along the entire camera ray. We contend this duplication introduces noise in empty and occluded spaces, posing challenges for producing high-quality 3D geometry. Drawing inspiration from traditional multi-view stereo methods, we propose an end-to-end 3D neural reconstruction framework CVRecon, designed to exploit the rich geometric embedding in the cost volumes to facilitate 3D geometric feature learning. Furthermore, we present Ray-contextual Compensated Cost Volume ($RCCV$), a novel 3D geometric feature representation that encodes view-dependent information with improved integrity and robustness. Through comprehensive experiments, we demonstrate that our approach significantly improves the reconstruction quality in various metrics and recovers clear fine details of the 3D geometries. Our extensive ablation studies provide insights into the development of effective 3D geometric feature learning schemes. Project page: \url{https://cvrecon.ziyue.cool}
\end{abstract}

\section{Introduction}
Monocular 3D reconstruction is a fundamental task in computer vision with wide-ranging applications, including augmented reality~\cite{neuralrecon,simplerecon}, virtual reality, robotics~\cite{colmap,robotic}, and autonomous driving\cite{monorec,advancing}. In recent years, learning-based methods~\cite{transformerfusion,atlas,simplerecon,vortx,neuralrecon} have shown promising results for this task. These methods use a sequence of posed images to predict a Truncated Signed Distance Field (TSDF) volume as the 3D reconstruction. They can be broadly categorized into two groups: volumetric-based and depth-based.

\begin{figure}
\centering
\includegraphics[width=\linewidth]{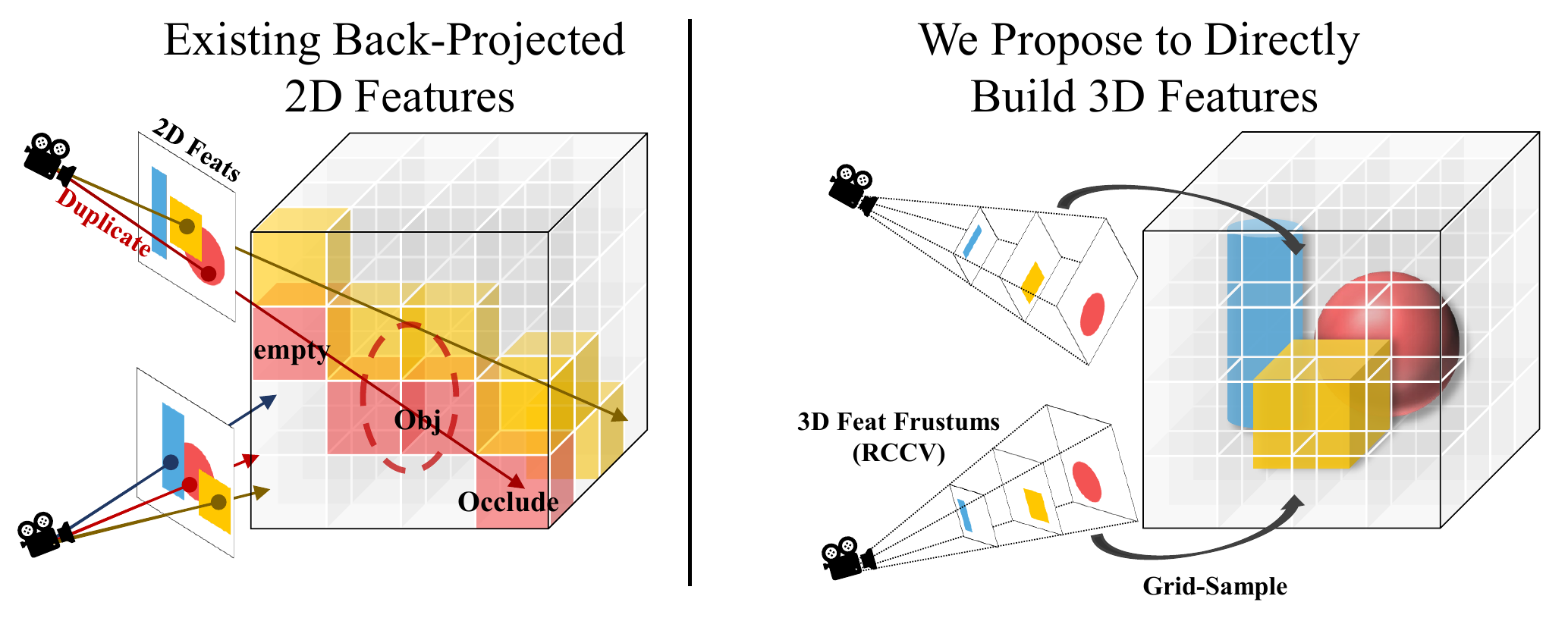}
\caption{\textbf{Novel 3D geometric feature learning paradigm.} Existing volumetric-based neural reconstruction methods simply back-project and duplicate the 2D image features along the entire camera ray, including empty spaces in front of the object and occluded areas behind the object, which will introduce noise and degrade the performance. We propose Ray-contextual Compensated Cost Volume ($RCCV$) as a novel 3D geometric feature representation. Geometries could be inferred from a fusion of the $RCCV$s.}
\label{fig:teaser}
\end{figure}

Existing volumetric-based methods~\cite{transformerfusion,atlas,vortx,neuralrecon} suffer from a critical limitation where the image feature is in the 2D while the reconstruction target is in 3D. As shown in Fig.~\ref{fig:teaser}, previous works simply duplicate the 2D features along the entire camera ray to the 3D space. When 2D image features representing object surfaces are placed in empty or occluded spaces without differentiation, it can complicate feature fusion and TSDF prediction in later stages, introduce noise and limit the model's ability to predict fine geometries. Therefore, it is more logical to directly build 3D feature representations that encode the geometry clue instead of simply filling it with 2D feature copies.

The cost volume, widely used in the depth prediction task, is a great choice for 3D geometric feature representation. It is composed of multi-view feature matching confidences across a set of pre-defined depth planes. A higher feature matching confidence at a given position indicates a greater likelihood of that position being an object surface. Compared to the simply duplicated 2D image features, cost volume explicitly encodes the depth probability distribution along the ray. However, existing depth-based reconstruction methods~\cite{pointbasedmvsnet,dpsnet,espnet,simplerecon,mvdepthnet,mvsnet} predict 2D depth maps from the 3D cost volume and then reconstruct the 3D structure through the TSDF Fusion~\cite{kinectfusion}. The 3D-2D-3D pipeline is not optimal because it doesn't take into account global context and frame consistency, which can result in the loss of structural details or the introduction of floating artifacts during the 3D-2D transformation. In contrast, if the 3D feature representation of each keyframe is retained and fused for the final reconstruction, the model will be able to estimate the structure holistically and unleash a greater potential.

In this paper, we propose to directly construct a novel Ray-contextual Compensated Cost Volume ($RCCV$) as a 3D geometric feature representation, which is a multi-view cost volume with 2 contributions: (1) For a keyframe image, we argue that the feature matching cost at a single location is not sufficient for inferring the geometry. As shown in Fig~\ref{fig:cvdist}, the confidence distribution along the entire camera ray is needed for reference: the highest matching confidence position within a camera ray is more likely to be the surface. Therefore, we devised a distribution feature for individual camera rays, which is subsequently integrated into each voxel along the said ray. (2) We observe the cost volume fails to encode any useful information in the non-overlapping regions and is excessively noisy in areas with low-contrasting textures as shown in Fig~\ref{fig:cost}. Therefore, besides the feature matching confidence, the original 2D image feature is also fused to the cost volume and we name it "Contextual Compensation". With these two contributions, our proposed $RCCV$ encodes comprehensive 3D geometric information. The reconstruction result could be inferred from a fusion of our $RCCV$s from multiple keyframes. Extensive experiments have demonstrated, as a generic representation, our $RCCV$ is agnostic to the downstream fusion and prediction models, and provides a more effective 3D geometric feature learning scheme.

To summarize, Our contributions are as follows:
\begin{itemize}
    \item We identify fundamental limitations of the existing feature learning scheme in the neural reconstruction field and accordingly propose to leverage the multi-view cost volume as a direct 3D geometric feature representation.
    \item We observe the widely-used standard cost volume lacks the distribution reference information along the camera ray and propose the Ray Compensation mechanism to solve this problem.
    \item To improve the robustness of the cost volume in the non-overlapping and low-texture areas, we propose a novel Contextual Compensation module.
    \item Our extensive experiments show the effectiveness of our proposed $RCCV$, and its agnostic nature with downstream fusion and prediction models.
\end{itemize}

\section{Related Works}
In this section, we first review relevant volumetric-based neural reconstruction methods and analyze their limitations. Then we briefly introduce the cost volume in the depth prediction field and survey its applications in the 3D reconstruction task.

\textbf{Volumetric-based 3D Reconstructions.}
In recent years 3D computer vision research~\cite{livepose,bridging,class,multimodal,pse} have shown remarkable progress, especially volumetric-based 3D reconstruction~\cite{transformerfusion,atlas,vortx,neuralrecon,finerecon}. These methods usually extract 2D image features and back-project them into the 3D feature volume, 3D dense or sparse convolutions are later applied to predict the TSDF volume from it. Finally, the 3D mesh can easily be obtained by marching cubes~\cite{marchingcubes}. Atlas~\cite{atlas} is the pioneering work that achieved promising results with simple back-projection and 3D convolution. NeuralRecon~\cite{neuralrecon} is later proposed to introduce a fragmenting strategy and RNN-based global fusion to handle large-scale environments. Moreover, VoRTX~\cite{vortx}, TransformerFusion~\cite{transformerfusion}, and concurrent work SDF-Former~\cite{sdfformer} explore the Transformer~\cite{transformer} mechanism for view selection, fusion, and aggregation, are complementary to our contribution.

Despite the progress made by volumetric-based methods, they suffer from fundamental limitations due to their feature representation. The duplications of image feature does not represent 3D geometry, inherently still a 2D feature. It also introduces noise by placing surface features in the front empty space and behind occluded areas, adding burdens to downstream models to estimate the 3D geometry, degrading the performance. In contrast, our novel $RCCV$ directly and explicitly encodes the 3D geometric information via cost volume representation (higher matching confidence indicates higher probability as object surface). 

\begin{figure*}
\centering
\includegraphics[width=.9\linewidth]{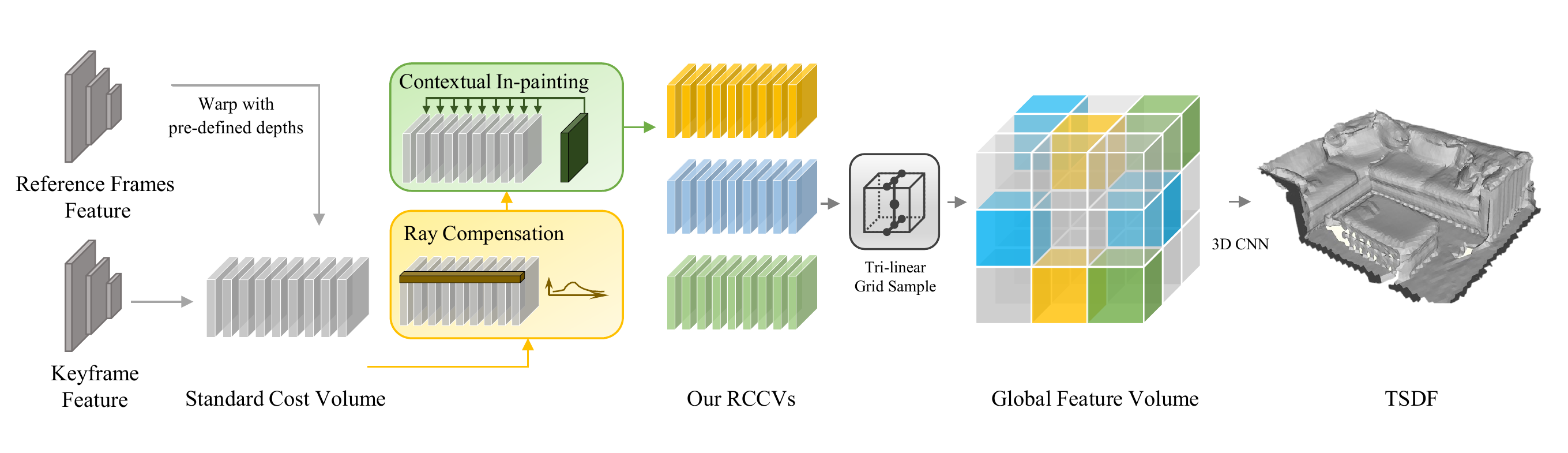}
\caption{\textbf{CVRecon Architecture:} We first build standard cost volumes for each keyframe with reference frames. Novel $RCCV$s are then generated with our proposed Ray Compensation and Contextual Compensation. We tri-linear grid-sample and fuse the $RCCV$s as a global feature volume and the TSDF reconstruction is later inferred with 3D Convolutions.}
\label{fig:framework}
\end{figure*}

\textbf{Depth-based 3D Reconstructions.}
The traditional 3D reconstruction methods typically involve predicting a dense depth map for each keyframe, which is then fused into a 3D structure. COLMAP~\cite{colmap} provides an impressive depth prediction baseline but suffers from low-texture regions. In the era of deep learning, neural network-based depth prediction models~\cite{pointbasedmvsnet,revisitingSI,dpsnet,espnet,mvdepthnet,mvsnet} usually build a cost volume to aid in the depth prediction. SimpleRecon~\cite{simplerecon} proposed to improve the robustness and accuracy by adding camera metadata to the cost volume. TSDF Fusion~\cite{kinectfusion}, RoutedFusion~\cite{routedfusion}, and NeuralFusion~\cite{neuralfusion} are proposed to fuse the predicted depth maps into a global TSDF representation.

However, depth-based methods have some essential limitations. First, the depth maps of different frames are predicted separately, leading to inevitable inconsistencies between the frames. Second, the depth representation only encodes the structure of the predicted 2D manifold, any imperfection in the depth prediction can cause to loss of actual surface geometry information from the cost volume. In contrast, our proposed method fuses all the geometric information of our novel $RCCV$s and predicts the 3D geometry holistically, ensuring consistency, accuracy, and completeness.

\textbf{Cost Volume in Depth Predictions.}
The cost volume is a successful geometric learning paradigm that has found wide application in flow estimation~\cite{camliflow,PWC-Net,wu2020pointpwc}, stereo matching~\cite{psmnet,gu2020cascade,lu2018sparse,mvsnet,wang2023real}, and depth prediction fields~\cite{dynamicdepth,depthformer,simplerecon,manydepth,yee2020fast}. In the depth scenario, multi-view RGB pixels or CNN features are projected onto pre-defined depth planes in the view-frustum, and pixel or feature matching costs are used to encode the depth probability distribution. The higher matching confidence regions in a camera ray are generally indicative of the object's surface. Existing methods flatten the 3D cost volume as a 2D feature map such that the cost distribution within each camera ray will be used to predict the depth value for the corresponding pixel. Conversely, we retain the 3D structure and introduce the Ray Compensation and Contextual Compensation to enable the costs in each 3D location to represent the geometry independently and robustly.

\section{Methodology}
Given a sequence of monocular images $\{I^i\}_{i=1}^M \in R^{3\times H\times W}$ with its 6-DOF poses $\{P^i\}_{i=1}^M\in SE(3)$ and intrinsics $\{K^i\}_{i=1}^M$, the goal of the 3D neural reconstruction is to predict a global TSDF volume $S$. Before the training, we use the TSDF Fusion~\cite{kinectfusion} to generate the ground-truth TSDF $S_{gt}$ from the ground truth depths. During the testing time, we do not have access to the ground truth depth.

\subsection{Method Overview}
The framework of our method is shown in Fig.~\ref{fig:framework}. To improve the efficiency, redundant image frames are removed and only $N$ keyframes $\{I^{i,0}\}_{i=1}^N$ are kept based on the pose distance. Each keyframe $I^{i,0}$ is associated with a set of reference frames $\{I^{i,k}\}_{k=1}^K$ with proper camera pose difference and view overlapping. We first build a standard cost volume $CV$ for each of the keyframes and then enhance them with our proposed Ray Compensation and Contextual Compensation. The generated $RCCV$s are integrated into a global feature volume through the process of grid sampling. Subsequently, a 3D CNN is employed to transform the volumetric representation into a TSDF volume.

Our key insights are as follows: (1) As shown in Fig.~\ref{fig:teaser}, we directly build $RCCV$ as a 3D geometric feature representation of the input image. Compare to the existing back-projection mechanism, our approach avoids introducing noise and improves the reconstruction quality. (2) We avoid the use of a 2D depth map as the intermediate representation, which suffers from inconsistency and would lose the information on the actual surface when the depth prediction is imperfect. Instead, we propose an end-to-end 3D framework CVRecon, which preserves all the geometric information to ensure an accurate holistic reconstruction. (3) We observe that the standard cost volume lacks the global context. As shown in Fig~\ref{fig:cvdist}, the cost distribution within a camera ray is not normalized and has multiple peaks. Predicting the geometry from a single cost value needs the ray distribution for reference. (4) The cost volume in the non-overlapping and texture-less areas does not carry much useful information as shown in Fig~\ref{fig:cost}. We accordingly propose Ray Compensation and Contextual Compensation to improve the integrity and robustness of the standard cost volume.

In the following sections, we first introduce the standard cost volume construction, followed by a detailed explanation of our proposed Ray Compensation and Contextual Compensation. Subsequently, the paper delves into the explication of the fusion mechanism, TSDF prediction, and loss function designs.

\subsection{Ray-contextual Compensated Cost Volume}
\label{sec:rccv}

\textbf{Standard Cost Volume Construction.} Consider an image keyframe $I^{t,0}$ with a set of reference frames $\{I^{t,k}\}_{k=1}^K$, cost volume is to encode the feature matching confidence at different 3D locations. The feature maps $\{F^{t,k}\}_{k=0}^K$ of all images are first extracted by a 2D CNN model called Matching Encoder $\theta_{ME}$ and then projected to a set of pre-defined depth hypothesis planes $\{D_i\}_{i=1}^{|D|}$. $|D|$ is the number of the depth hypothesis. For the keyframe $I^{t,0}$, the $i$th plane $CV^t_i$ in the standard cost volume $CV^t$ is the concatenation of the dot products of the projected reference frame features $\{ \hat{F}^{t,k}_i \}_{k=1}^K $ and the keyframe feature $F^{t,0}$. $\hat{F}_i$ indicates the feature $F$ is perspective projected with depth $D_i$.

\begin{eqnarray}
F^{t,k} &=& \theta_{ME}(I^{t,k}), \\
\hat{F}^{t,k}_i &=& \pi_0(\pi^{-1}_k(F^{t,k}, D_i)), \\
CV^t_i &=& \left \langle F^{t,0} \cdot  \hat{F}^{t,k}_i \right \rangle _{k=1}^K , \\
CV^t &=& \left \langle CV^t_i \right \rangle^{|D|}_{i=1},
\end{eqnarray}
where $\pi$ is the perspective projection function based on the corresponding intrinsic and pose, and $\langle \rangle$ is the concatenation operation.

\begin{figure}
\centering
\includegraphics[width=.8\linewidth]{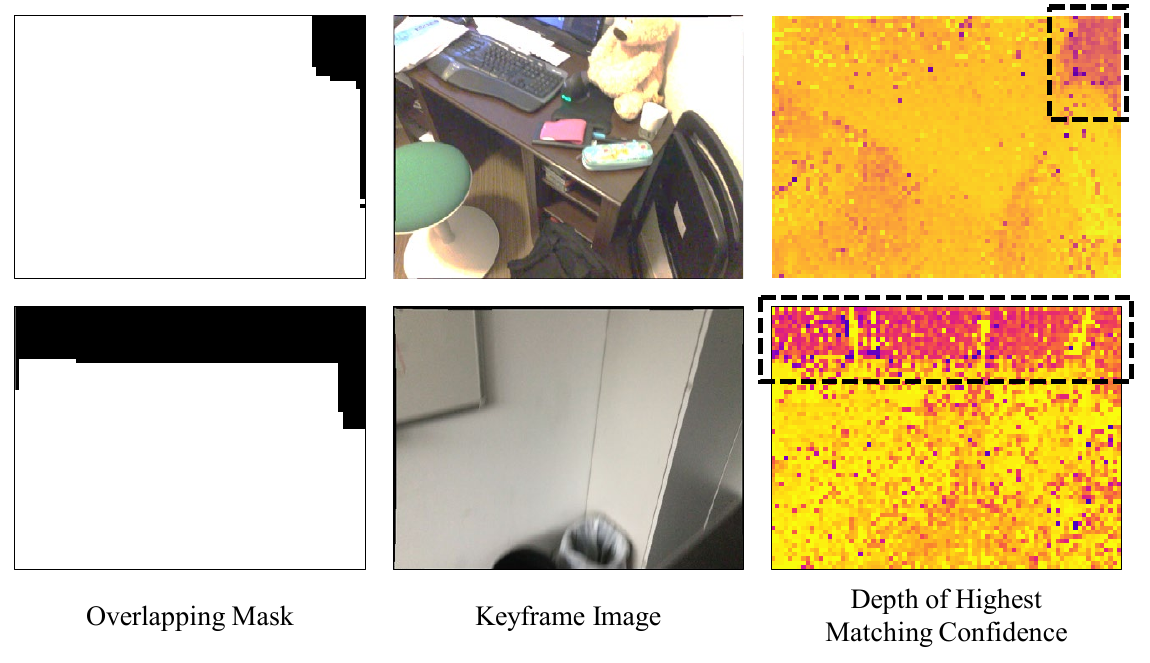}
\caption{\textbf{Standard Cost Volumes are Noisy in Non-overlapping and Low-texture Areas.} The black masks and dashed rectangular markers indicate there is no reference frame overlap with the keyframe. We propose Contextual Compensation to improve the robustness in the non-overlapping and low-texture areas.}
\label{fig:cost}
\end{figure}

\begin{figure}
\centering
\includegraphics[width=.9\linewidth]{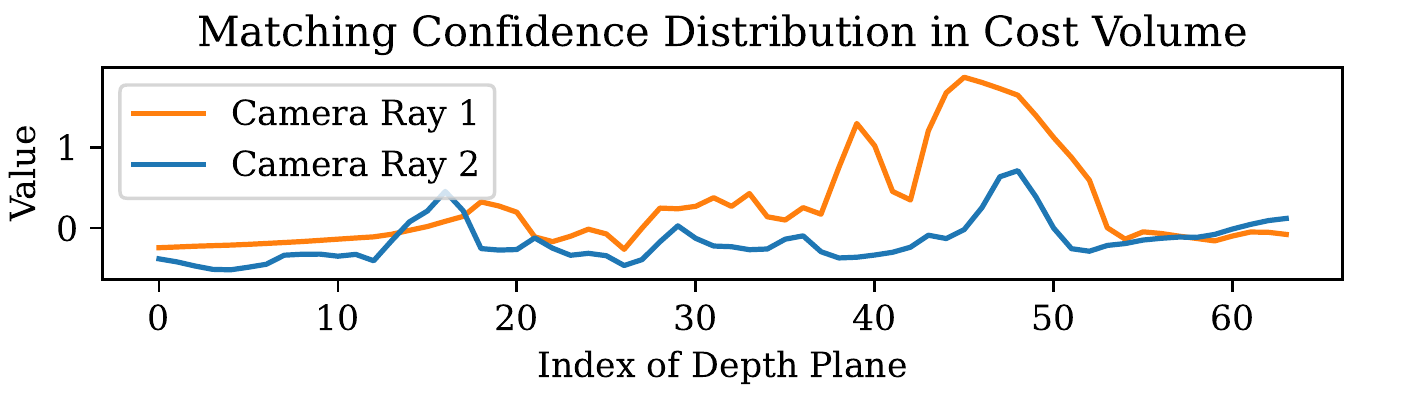}
\caption{\textbf{Camera Ray Distributions in Cost Volume.} We visualize a channel of matching confidence distributions of two sample camera rays in the cost volume. These distributions have multiple peaks and the values are not normalized in magnitude. The model tends to decode local maxima as the object surfaces without the ray distribution as a reference, causing floating artifacts as shown in Fig~\ref{fig:floating}.}
\label{fig:cvdist}
\end{figure}

\begin{figure}
\centering
\includegraphics[width=.9\linewidth]{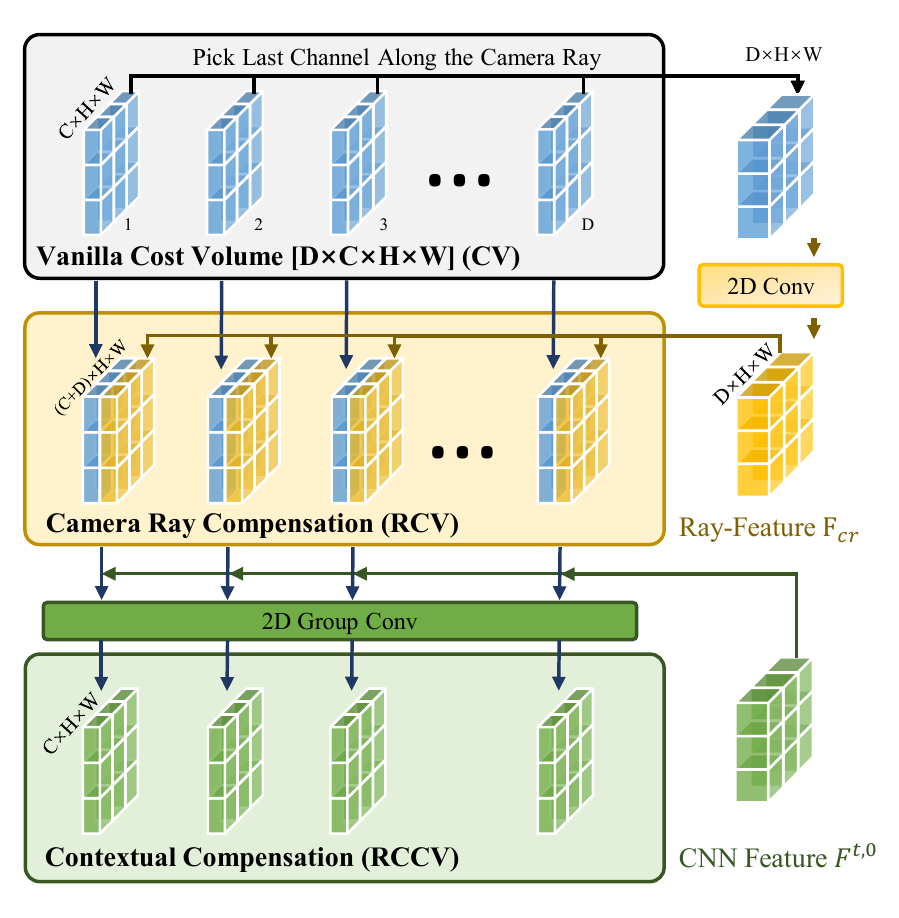}
\caption{\textbf{Ray Compensation and Contextual Compensation.} Decoding the geometry from the cost value requires the overall distribution information within its camera ray for reference. We propose Ray Compensation to concatenate the ray information to all depth planes of the cost volume. The matching cost is too noisy in the non-overlapping and low-texture areas, the Contextual Compensation is proposed to fuse 2D CNN features to improve the robustness.}
\label{fig:raycompensation}
\end{figure}

\textbf{Camera Ray Compensation.} We observe that directly using the above standard cost volume as 3D geometric feature does not shown significant performance improvement because it lacks camera ray information. Specifically, for each keyframe $I^{t,0}$, we build a standard cost volume $CV^t \in R^{|D| \times C \times H \times W}$ where $C$ is the channel number and $H, W$ are widths and heights in the image plane. The camera ray $R^{|D| \times C}$ of each pixel $[h, w]$ encodes its $C$ channels depth probability distribution as shown in Fig~\ref{fig:cvdist}. This standard cost volume is view-dependent, whether a cost value represents an object surface is conditioned on all front cost values in its camera ray. Conventional depth prediction methods reshape this standard cost volume as a 2D feature map $R^{(|D|\cdot C)\times H \times W}$ and predict each pixel's depth with its entire camera ray information. However, for 3D reconstruction we expect a view-independent 3D feature representation. If we directly employ the standard cost volume as a 3D feature representation, the single feature $R^C$ of a 3D location $[d, h, w]$ solely is not sufficient to decode its geometry without the overall camera ray distribution. Based on this observation, as shown in Fig~\ref{fig:raycompensation}, we propose a Camera Ray Compensation module that builds a camera ray feature $F^t_{cr}$ and concatenates it into this standard cost volume $CV^t$ to form our Ray-compensated Cost Volume $RCV^t$.

\newcommand{\x}{ x }
\newcommand{\midline}{  }

\newcommand{\splitline}{\arrayrulecolor{black}\hhline{~------------}}

\newcommand*\rot{\rotatebox{90}}

\definecolor{Asectioncolor}{RGB}{255, 200, 200}
\definecolor{Bsectioncolor}{RGB}{255, 228, 196}
\definecolor{Csectioncolor}{RGB}{235, 255, 235}
\definecolor{Dsectioncolor}{RGB}{235, 235, 255}
\newcolumntype{a}{>{\columncolor{Gray}}c}
\newcommand{\first}{\cellcolor{firstcolor}\textbf}
\newcommand{\second}{\cellcolor{secondcolor}\underline}

\begin{table*}[t]
  \centering
  \resizebox{.9\textwidth}{!}{
    \begin{tabular}{|c|l|c||c|c|c||c|c|a|}
        \arrayrulecolor{black}\hline
          \multirow{2}*{Eval} &\multirow{2}*{Method} &\multirow{2}*{3D Feature Source}  &\multicolumn{3}{c||}{\cellcolor{col1}The lower the better} &\multicolumn{3}{c|}{\cellcolor{col2}The higher the better}\\
          ~& ~ &~ &\cellcolor{col1}Accuracy & \cellcolor{col1}Completeness & \cellcolor{col1}Chamfer  & \cellcolor{col2}Precision & \cellcolor{col2}Recall & \cellcolor{col2}F1 Score @ 5cm \\

        \hline\hline
        \parbox[b]{2mm}{\multirow{7}{*}{\rotatebox[origin=c]{90}{Atlas~\cite{atlas}}}} 
        & DeepVideoMVS~\cite{dvmvs}                         & 2D Depth      & 0.079             & 0.133             & 0.106             & 0.521             & 0.454             & 0.474 \\ \midline

        & SimpleRecon~\cite{simplerecon}                    & 2D Depth      & 0.065             & \second{0.078}    & \second{0.071}    & 0.641             & 0.581             & 0.608 \\ \midline
        
        & NeuralRecon~\cite{neuralrecon}                    & Proj 2D feats     & 0.054             & 0.128             & 0.091             & 0.684             & 0.479             & 0.562 \\ \midline
        
        & Atlas~\cite{atlas}                                & Proj 2D feats     & 0.068             & 0.098             & 0.083             & 0.640             & 0.539             & 0.583 \\ \midline

        & TransformerFusion~\cite{transformerfusion}        & Proj 2D feats     & 0.078             & 0.099             & 0.088             & 0.648             & 0.547             & 0.591 \\ \midline
        
        & VoRTX~\cite{vortx}                                & Proj 2D feats     & \second{0.054}    & 0.090             & 0.072             & \second{0.708}    & \second{0.588}    & \second{0.641} \\ \midline
        
        & \textbf{Ours}                                  &\textbf{3D RCCV} & \first{0.045}     & \first{0.077}     & \first{0.061}     & \first{0.753}     & \first{0.639}     & \first{0.690} \\
        \midline
        
        \hline\hline
        \parbox[b]{2mm}{\multirow{11}{*}{\rotatebox[origin=c]{90}{TransformerFusion~\cite{transformerfusion}}}} 
        & COLMAP~\cite{colmap}                              & 2D Depth      & 0.102             & 0.118             & 0.110             & 0.509             & 0.474             & 0.489 \\ \midline
        
        & DPSNet~\cite{dpsnet}                              & 2D Depth      & 0.119             & 0.075             & 0.097             & 0.474             & 0.519             & 0.492 \\ \midline
        
        & DELTAS~\cite{deltas}                              & 2D Depth      & 0.119             & 0.074             & 0.097             & 0.478             & 0.533             & 0.501 \\ \midline
        
        & DeepVideoMVS~\cite{dvmvs}                         & 2D Depth      & 0.106             & 0.069             & 0.087             & 0.541             & 0.592             & 0.563 \\ \midline

        & 3DVNet~\cite{3dvnet}                              & 2D Depth      & 0.077             & 0.067             & 0.072             & 0.655             & 0.596             & 0.621 \\ \midline
        
        & SimpleRecon~\cite{simplerecon}                    & 2D Depth      & 0.055             & \first{0.060}     & 0.058             & 0.686             & \second{0.658}    & 0.671 \\ \midline
        
        & NeuralRecon~\cite{neuralrecon}                    & Proj 2D feats     & 0.051             & 0.091             & 0.071             & 0.630             & 0.612             & 0.619 \\ \midline
        
        & Atlas~\cite{atlas}                                & Proj 2D feats     & 0.072             & 0.076             & 0.074             & 0.675             & 0.605             & 0.636 \\ \midline
        
        & TransformerFusion~\cite{transformerfusion}        & Proj 2D feats     & 0.055             & 0.083             & 0.069             & 0.728             & 0.600             & 0.655 \\ \midline
        
        & VoRTX~\cite{vortx}                                & Proj 2D feats     & \second{0.043}    & 0.072             & \second{0.057}    & \second{0.767}    & 0.651             & \second{0.703} \\ \midline
        
        & \textbf{Our CVRecon}                              &\textbf{3D RCCV} & \first{0.038}     & \second{0.067}    & \first{0.053}     & \first{0.794}     & \first{0.685}     & \first{0.735} \\
        \midline
         
        \arrayrulecolor{black}\hline

    \end{tabular}
  } 
  \vspace{1mm}
  \caption{\textbf{3D Mesh Evaluation on ScanNet2.} The upper part follows the evaluation metric from Atlas~\cite{atlas} while the lower part follows TransformerFusion~\cite{transformerfusion}. Methods in each category are sorted by the F1 Score. The \colorbox{firstcolor}{\textbf{best}} and \colorbox{secondcolor}{\underline{second-best}} scores are marked respectively. Existing volumetric methods simply duplicate 2D image features to the 3D space, depth-based methods predict 2D depth maps and use TSDF-Fusion to retrieve the 3D geometry. In contrast, our end-to-end 3D framework CVRecon with novel 3D geometric feature representation $RCCV$ is a more natural and logical design.
    }

\label{tab:mesh}
\end{table*}

\begin{eqnarray}
F^t_{cr} &=& \text{Conv2d}( \left \langle CV^t_i[-1]\right \rangle _{i=1}^{|D|} ),\\
RCV^t_i &=& \langle CV^t_i, F^t_{cr}\rangle ,\qquad \ \ ,  \\
\quad RCV^t &=& \langle RCV^t_i\rangle_{i=1}^{|D|}
\end{eqnarray}
where $[-1]$ is the tensor indexing operation that picks the last element of the channel dimension. 

Before the ray compensation, the channel number of standard cost volume is reduced by a small MLP to $C=7$. Each of the $C$ channels is an aggregation of all information from all frames. We find that using all channels to construct the ray context feature only increases $0.002$ of the F1-score but consumes $7$ times memory. We find that only picking one channel performs well enough and which channel to pick does not matter.

\textbf{Contextual Compensation.} As shown in Fig~\ref{fig:cost}, we still observe some fundamental limitations of our Ray-compensated Cost Volume $RCV$: (1)When the camera is moving backward or rotating, some areas in the new image frame may not overlap with the previous frames; (2)Some objects like floors and walls have very low contrast in textures. In these areas, the feature matching cost is noisy and could not provide reliable geometric information. To solve these problems, as shown in Fig~\ref{fig:raycompensation}, we propose a Contextual Compensation module to fuse the 2D CNN feature $F^{t,0}$ of the keyframe $t$ into the $RCV^t$ to form the Ray-contextual Compensated Cost Volume $RCCV^t$. In the non-overlapping and low-texture areas, instead of fully noise in standard cost volume, our $RCCV$ will degrade and similar to duplicated 2D CNN features in the existing works, improving the robustness. We observe that using separate convolution kernel weights for each depth plane generates significantly better results and we hypothesize it is related to the different spatial scales across depth planes. We efficiently implement the Contextual Compensation by concatenating the cost volume feature and 2D feature, followed by a group convolution operation with the $|D|$ as the group number.

\begin{figure}
\centering
\includegraphics[width=.85\linewidth]{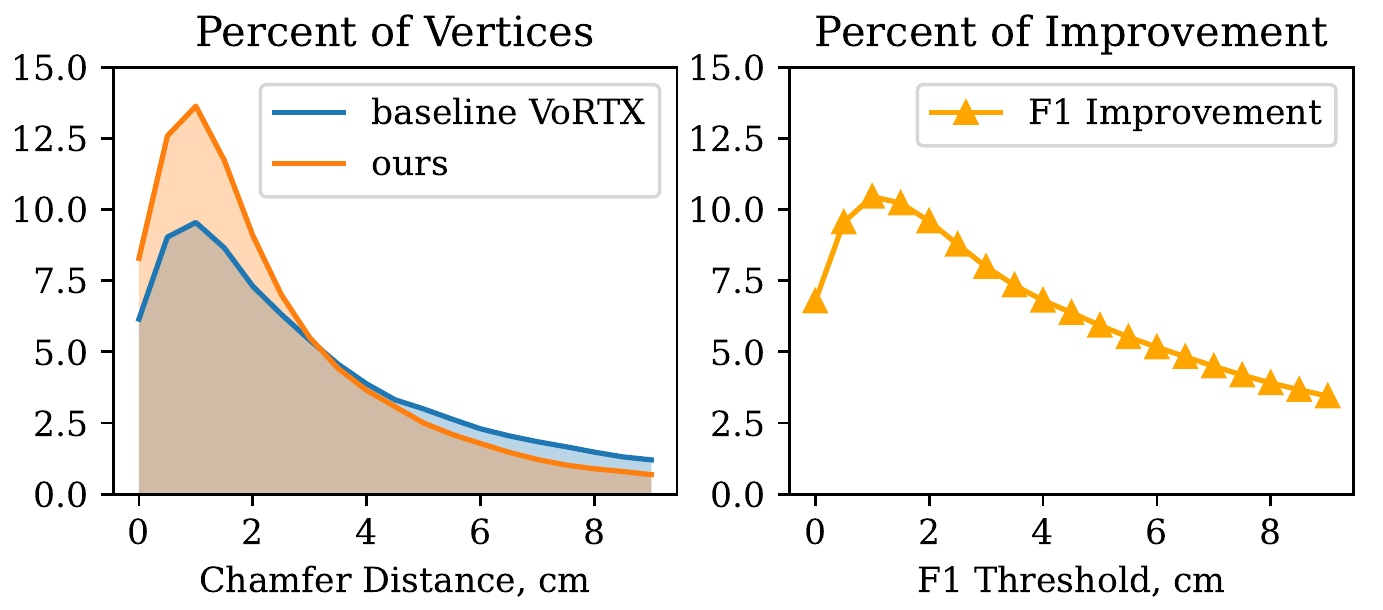}
\caption{\textbf{Analyze of 3D Mesh Improvements.} The left figure is the chamfer distance distribution between the predicted and ground-truth meshes, which is generally lower in our predictions. The right figure shows our improvement percentage of the F1 score with different evaluation thresholds, the widely used 5cm F1 score could not fully capture our advantage.}
\label{fig:threshold}
\end{figure}

\definecolor{Asectioncolor}{RGB}{255, 200, 200}
\definecolor{Bsectioncolor}{RGB}{255, 228, 196}
\definecolor{Csectioncolor}{RGB}{235, 255, 235}
\definecolor{Dsectioncolor}{RGB}{235, 235, 255}


\begin{table*}[t]
  \centering
  \resizebox{.9\textwidth}{!}{
    \begin{tabular}{|c|l|c||a|c|c|c|c||c|c|c|}
        \arrayrulecolor{black}\hline
          \multirow{2}*{} &\multirow{2}*{Method} &\multirow{2}*{Depth-Supervised}  &\multicolumn{5}{c||}{\cellcolor{col1}The lower the better} &\multicolumn{3}{c|}{\cellcolor{col2}The higher the better}\\
          ~& ~ &~ &\cellcolor{col1}Abs Diff &\cellcolor{col1}Abs Rel &\cellcolor{col1}Sq Rel & \cellcolor{col1}RMSE & \cellcolor{col1}logRMSE  & \cellcolor{col2}$\delta < 1.25 $ & \cellcolor{col2}$\delta < 1.25^{2}$ & \cellcolor{col2}$\delta < 1.25^{3}$ \\

        \hline\hline
        \parbox[b]{2mm}{\multirow{6}{*}{\rotatebox[origin=c]{90}{Depth-based}}} 

        & COLMAP~\cite{colmap}                              & Yes       & 0.264         & 0.137        & 0.138        & 0.502        & -            & 83.4         & -           & -         \\ \midline
        
        & GPMVS~\cite{gpmvs}                                & Yes       & 0.239         & 0.130        & 0.339        & 0.472        & 0.108        & 90.6         & 96.7        & 98.0         \\ \midline

        & MVDepthNet~\cite{mvdepthnet}                      & Yes       & 0.191         & 0.098        & 0.061        & 0.293        & 0.116        & 89.6         & 97.7        & 99.4         \\ \midline
        
        & DPSNet~\cite{dpsnet}                              & Yes       & 0.158         & 0.087        & 0.035        & 0.232        & 0.110        & 92.5         & 98.4        & 99.5         \\ \midline
        
        & DELTAS~\cite{deltas}                              & Yes       & 0.149         & 0.078        & 0.027        & 0.221        & 0.107        & 93.7         & -           & -         \\ \midline

        & SimpleRecon~\cite{simplerecon}                    & Yes       & 0.088         & 0.043        & 0.012        & 0.146        & 0.067        & 98.0         & -           & -         \\ \midline
        
        \hline\hline
        \parbox[b]{2mm}{\multirow{4}{*}{\rotatebox[origin=c]{90}{Volumetric}}} 
        
        & Atlas~\cite{atlas}                                & No        & 0.124         & 0.065        & 0.043        & 0.251        & -            & 93.6         & 97.1        & 98.6         \\ \midline
        
        & NeuralRecon~\cite{neuralrecon}                    & No        & 0.106         & 0.065        & 0.031        & 0.195        & -            & 94.8         & 96.1        & 97.5         \\ \midline
        
        & VoRTX~\cite{vortx}                                & No        & 0.096         & 0.061        & 0.038        & 0.205        & -            & 94.3         & 97.3        & 98.7         \\ \midline
        
        & \textbf{Our CVRecon}                              & No        &\textbf{0.078} &\textbf{0.047}&\textbf{0.028}&\textbf{0.181}&\textbf{0.094}&\textbf{96.3} &\textbf{98.2}&\textbf{99.1}  \\ \midline
        \midline
         
        \arrayrulecolor{black}\hline

    \end{tabular}
  } 
  \vspace{1mm}
  \caption{\textbf{2D Depth Evaluation on ScanNet2.} The upper parts are depth-based methods, which predict depths and are directly supervised by the ground-truth depths. The lower parts are volumetric methods supervised by the ground-truth TSDF, we render the depth maps from its mesh predictions. All methods are ranked by the absolute depth difference within their category, best scores are in \textbf{bold}.
    }

\label{tab:depth}
\end{table*}

\begin{eqnarray}
RCCV^t_i &=& \text{Conv2d}_i(\langle RCV^t_i, F^{t,0}\rangle),  \\
RCCV &=& \langle RCCV_i\rangle _{i=1}^{|D|}
\end{eqnarray}

\textbf{Fusion of the RCCVs.} After obtaining the $RCCV$ for each image keyframe, we generate a global feature volume by grid sampling with tri-linear interpolation. Given the downstream operation-agnostic nature of our proposed RCCV feature, we have found that it can be seamlessly integrated with various inter-frame feature fusion techniques, such as the multi-head self-attention module~\cite{vortx} or naive averaging operation ~\cite{atlas,neuralrecon}.

\textbf{TSDF Prediction.} We employ 3D dense~\cite{atlas} or sparse~\cite{vortx,neuralrecon} convolution modules for geometry prediction. The predictions at the coarse and medium levels are occupancy grids to sparsify the feature grid while at the fine level, we directly predict the TSDF volume.

\textbf{Loss Functions.} Following NeuralRecon~\cite{neuralrecon}, we apply binary cross-entropy (BCE) loss function to the coarse and medium level occupancy predictions and $l1$ loss function to the fine level TSDF prediction. The TSDF ground truth is in $4\ cm$ resolution. Following Atlas~\cite{atlas}, we mark all unobserved columns of the ground truth TSDF volume as unoccupied.

\subsection{Implementation Details}
For the cost volume, we set the depth range from 0.25 to 5.0 meters and divide it into 64 depth planes evenly in the log space. We construct the cost volume at $1/8 (80\times60)$ of the input image resolution $(640\times 480)$. Following SimpleRecon~\cite{simplerecon}, we concatenate metadata (rays, angles, pose, etc..) into the cost volume and use an MLP to reduce the channel dimension to 7. In Ray Compensation, the channel number of our ray feature $F_{cr}$ is the same as $|D|$ which is 64. After the Contextual Compensation, our group convolution reduces the channel number back to 7. The final $RCCV$ is a tensor of $R^{64 \times 7 \times 60 \times 80}$ which consumes around $4MB$ of memory in the FP16 precision and $15ms$ computation time.

Our proposed $RCCV$ is agnostic to the downstream fusion and prediction models which consume most of the computation power. When applying our $RCCV$ to the VoRTX~\cite{vortx} or Atlas~\cite{atlas} downstream models, the training takes around $40-45$ hours on two Nvidia A100 GPUs, showing minor computational overhead.

\section{Experiments}

\begin{figure}
\centering
\includegraphics[width=.8\linewidth]{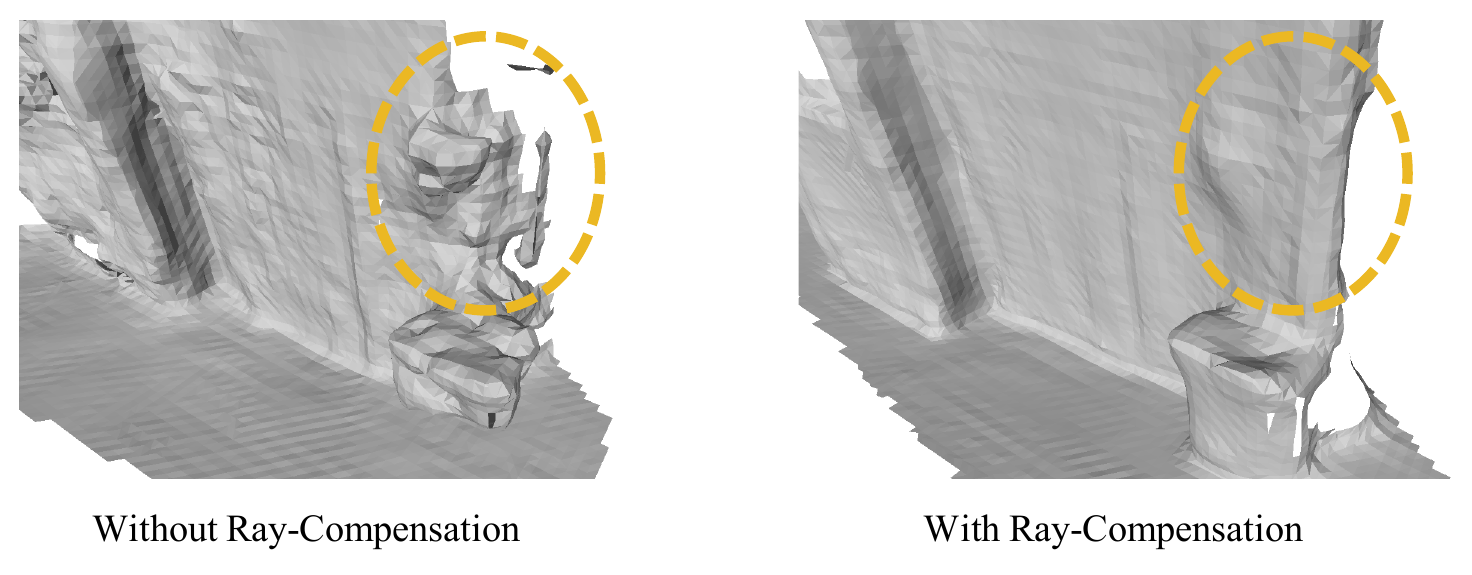}
\caption{\textbf{Floating Artifacts without Ray-compensation.} Without the ray distribution as a reference, the model tends to decode local maxima of the matching confidence in the cost volume as object surfaces, causing floating artifacts.}
\label{fig:floating}
\end{figure}

To evaluate the effectiveness of our new 3D geometric feature representation $RCCV$, we apply it to a volumetric 3D reconstruction baseline VoRTX~\cite{vortx} to replace the simple back-projection mechanism and name it CVRecon. We evaluate its performance on the challenging ScanNet2~\cite{scannet} dataset, which comprises 1513 RGB-D scans from 707 indoor spaces. Following the official split, our training, validation, and testing sets consist of 1201, 312, and 100 scans, respectively.

In the following sections, we evaluate our reconstruction quality with 3D mesh metrics and 2D depth metrics. Then we conduct a comprehensive ablation study to dig into each part of our contributions to analyze their qualitative and quantitative impact. We also adopt our $RCCV$ to different baselines to verify it is agnostic to downstream modules.

\begin{table*}[t]
\centering
\resizebox{.9\textwidth}{!}{
\begin{tabular}{|c|c|c|c|c|c|cccc|}
\hline
\cellcolor{col2}Back-Project &\cellcolor{col2}Grid Sample &\cellcolor{col2}Camera Ray &\multicolumn{3}{c|}{\cellcolor{col2}Contextual Inpainting} &\multicolumn{4}{c|}{\cellcolor{col1}3D Mesh Metrics}\\
\cellcolor{col2}2D Image Feats &\cellcolor{col2} Cost Volume&\cellcolor{col2}Compensation &\cellcolor{col2}Concat&\cellcolor{col2}Uni$-$Conv &\cellcolor{col2}Group$-$Conv &\cellcolor{col1}Acc $\downarrow$ &\cellcolor{col1}Comp $\downarrow$ & \cellcolor{col1}Chamfer $\downarrow$ & \cellcolor{col1}F1-Score $\uparrow$ \\
\hline
\multicolumn{10}{|c|}{\cellcolor{Gray}Evaluating 3D Feature Construction}\\
\hline
\checkmark &           &           &           &           &             &0.043   &   0.072  &   0.057  &   0.703 \\
\hline
           &\checkmark &           &           &           &             &0.041   &   0.071  &   0.056  &   0.711\\
\hline 
\multicolumn{10}{|c|}{\cellcolor{Gray}Evaluating Camera Ray Compensation}\\
\hline
           &\checkmark &           &           &           &             &0.041   &   0.071  &   0.056  &   0.711\\
\hline 
           &\checkmark &\checkmark &           &           &             &0.040   &   0.068  &   0.054  &   0.728\\
\hline
\multicolumn{10}{|c|}{\cellcolor{Gray}Evaluating Contextual Inpainting}\\
\hline
           &\checkmark &\checkmark &           &           &             &0.040   &   0.068  &   0.054  &   0.728\\
\hline
           &\checkmark &\checkmark &\checkmark &           &             &0.039   &   0.067  &   0.053  &   0.730\\
\hline
           &\checkmark &\checkmark &\checkmark &\checkmark &             &0.038   &   0.067  &   0.053  &   0.731\\
\hline
           &\checkmark &\checkmark &\checkmark &           &\checkmark   &\textbf{0.038} &\textbf{0.067} &\textbf{0.053} &\textbf{0.735}\\
\hline
\end{tabular}}
\vspace{1mm}
\caption{\textbf{Ablation Study: }
Evaluating the effectiveness of our proposed novel 3D geometric feature learning scheme, including the cost volume, Camera Ray Compensation, and Contextual In-painting on the ScanNet2~\cite{scannet} dataset with the evaluation protocol from the TransformerFusion~\cite{transformerfusion}.
}
\label{tab:ablation}
\end{table*}

\subsection{3D Reconstruction Evaluations}

The quantitative 3D mesh evaluation results on the ScanNet2~\cite{scannet} dataset are shown in Table~\ref{tab:mesh}. We evaluate our method with the standard protocol proposed by Atlas~\cite{atlas} and a more reasonable protocol proposed by TransformerFusion~\cite{transformerfusion} which excluded the unobserved areas. The ground-truth meshes are incomplete at the unseen areas in the input monocular sequences while many learning-based methods are able to hallucinate those structures. The TransformerFusion~\cite{transformerfusion} evaluation protocol uses an occlusion mask to prevent penalizing the more complete reconstruction.

Our CVRecon with the novel $RCCV$ outperforms a wide range of state-of-the-art reconstruction methods by a large margin. The depth-based methods~\cite{dvmvs,dpsnet,3dvnet,simplerecon,colmap,deltas} use predicted 2D depth maps as the intermediate representation and retrieve 3D structure by the TSDF Fusion~\cite{kinectfusion}, these methods suffer from (1) inconsistent depth prediction and (2) potentially lost information of the actual surface whenever the depth prediction is imperfect. Volumetric-based methods simply back-project 2D image features to all voxels along the camera ray, introducing noise to empty and occluded spaces. In contrast, our novel 3D feature representation $RCCV$ and the end-to-end 3D framework CVRecon solve these problems and significantly improve the reconstruction quality.

\begin{figure*}
\centering
\includegraphics[width=.9\linewidth]{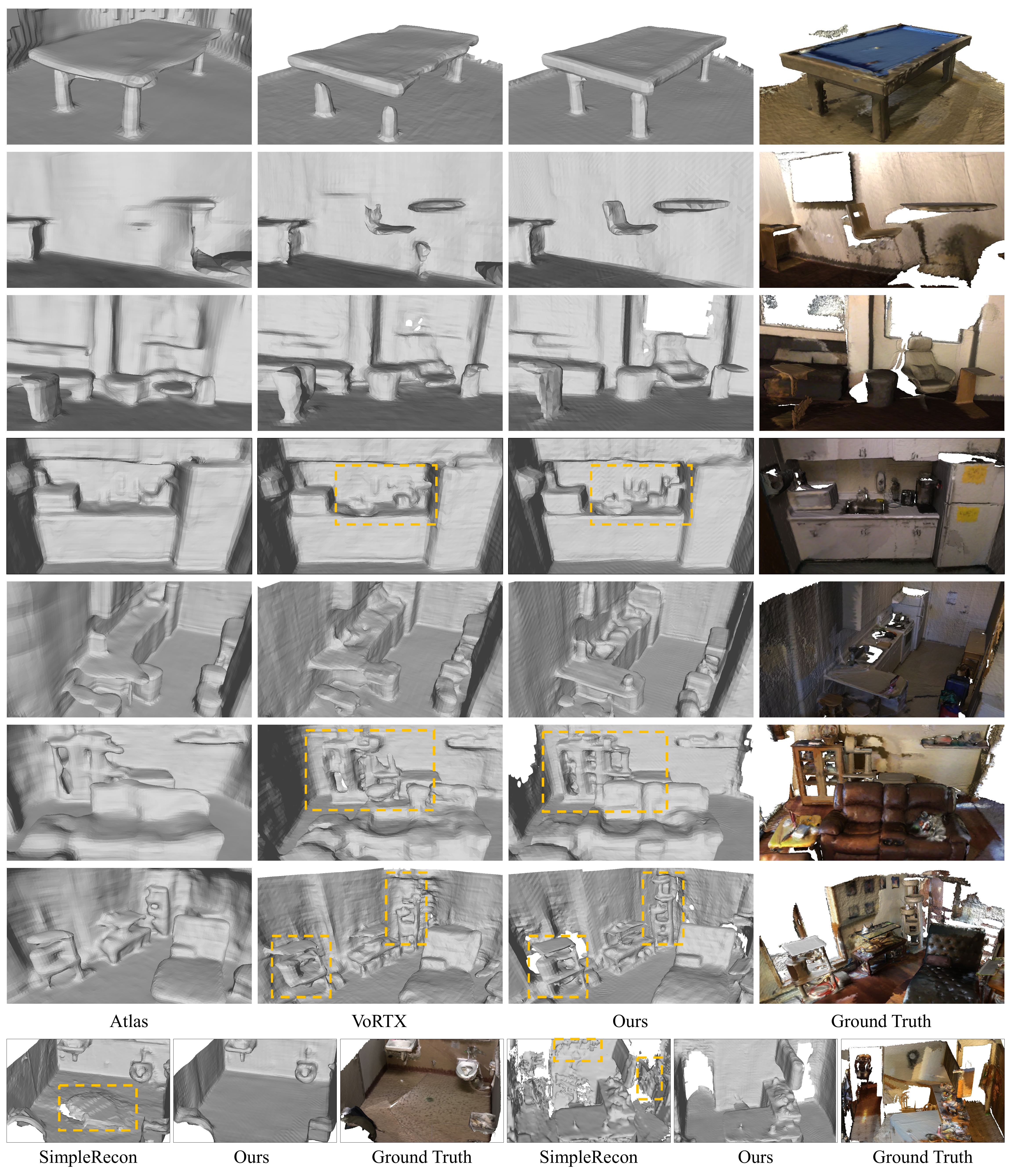} 
\caption{\textbf{Qualitative Comparison:} In the above part we compare our method with Atlas~\cite{atlas} and the current state-of-the-art VoRTX~\cite{vortx} on the ScanNet2~\cite{scannet} dataset test set. Note that the only difference with the VoRTX~\cite{vortx} is we use our $RCCV$ as the 3D geometric feature representation, which leads to significantly clear geometry details. We compare our method with the state-of-the-art depth-based method SimpleRecon~\cite{simplerecon} in the last row. The lack of translation parallax in narrow spaces like the left sample and the faraway texture-less walls in the right sample will lead to severe inconsistency and degrade the performance of the depth-based methods. In contrast, our holistic prediction generates a much clear and smooth reconstruction. More samples and analysis are in the supplementary materials.}
\label{fig:qualitative}
\end{figure*}

Qualitative visualizations are shown in Fig~\ref{fig:qualitative}. Compared to existing volumetric methods~\cite{atlas,vortx}, our CVRecon generates much more clear and more complete fine details. Our proposed novel 3D geometric feature representation $RCCV$ significantly unlocks the potential of downstream models for better reconstruction quality. Compared to the state-of-the-art depth-based model SimpleRecon~\cite{simplerecon}, one of the major advantages of our method is to holistically generate more consistent and clean geometries. More analysis could be found in the supplementary materials.

Moreover, we would like to point out that the default 5 cm threshold for F1-score is excessively large and fails to adequately capture our superior performance in modeling fine details of the geometry. As shown in Fig~\ref{fig:threshold}, our method reduces a significant portion of mesh vertex chamfer distances to under 2 cm, smaller thresholds demonstrate much higher improvements.

\subsection{2D Depth Evaluations}
We evaluate the 2D depth metrics on the ScanNet2~\cite{scannet} dataset and compare them with existing state-of-the-art methods in Table~\ref{tab:depth}. Depth-based reconstruction methods are supervised by ground-truth depths. Volumetric methods like ours directly predict the 3D geometry and are supervised by the ground-truth TSDF volume. We render pseudo-depths from predicted meshes to perform this evaluation for volumetric methods.

Our CVRecon significantly outperforms all other volumetric methods on all metrics by a large margin. We even outperform all the depth-based methods on the absolute difference metric. We attribute this performance boost to the rich geometric encoding in our $RCCV$. Another interesting observation is that depth-based methods generally perform better on relative metrics but worse on absolute metrics because they focused too much on the near objects.

\subsection{Ablation Study}

In this section, we conduct extensive ablation studies on the ScanNet2~\cite{scannet} dataset to comprehensively evaluate the effectiveness of our proposed novel 3D geometric feature learning scheme. As shown in Table~\ref{tab:ablation}, we perform 3 groups of experiments to evaluate the standard cost volume, Ray Compensation, and Contextual Compensation.

\textbf{Cost Volume as 3D Feature Representation.} In the first group of our experiment, we compare the simple back-projection of the 2D image feature and the grid sampling of the standard cost volume. The result confirms that 3D reconstruction quality could be improved by not polluting the feature volume of the empty and occluded areas, as well as the rich 3D geometric encoding in the standard cost volume.

\definecolor{Asectioncolor}{RGB}{255, 200, 200}
\definecolor{Bsectioncolor}{RGB}{255, 228, 196}
\definecolor{Csectioncolor}{RGB}{235, 255, 235}
\definecolor{Dsectioncolor}{RGB}{235, 235, 255}

\definecolor{darkgreen}{rgb}{0, 0.7, 0}
\newcommand{\greenbf}[1]{\textcolor{darkgreen}{\bf #1}}
\newcommand{\green}[1]{\textcolor{darkgreen}{#1}}

\begin{table}[t]
  \centering
  \resizebox{.48\textwidth}{!}{
    \begin{tabular}{|c|c|c|c|c|c|c|}
        \arrayrulecolor{black}\hline
        \multirow{2}*{Method}  &\multicolumn{3}{c|}{\cellcolor{col1}The lower the better} &\multicolumn{3}{c|}{\cellcolor{col2}The higher the better}\\
        ~ &\cellcolor{col1}Accuracy & \cellcolor{col1}Completeness & \cellcolor{col1}Chamfer  & \cellcolor{col2}Precision & \cellcolor{col2}Recall & \cellcolor{col2}F1 Score \\

        \hline
    
         Atlas~\cite{atlas} + Back-proj                             & 0.076             & 0.085             & 0.081             &0.644              & 0.559             & 0.596         \\ \midline
    
         Atlas~\cite{atlas} + RCCV                     & \textbf{0.070}     & \textbf{0.073}     & \textbf{0.071}     & \textbf{0.683}     & \textbf{0.620}     & \textbf{0.648} \\

         \hline

         \greenbf{Improvement}                            & \greenbf{7.9\%}     & \greenbf{14.1\%}     & \greenbf{12.3\%}     & \greenbf{6.1\%}     & \greenbf{10.9\%}     & \greenbf{8.7\%} \\
        
        \arrayrulecolor{black}\hline

    \end{tabular}
  } 
  \vspace{1mm}
  \caption{ \textbf{Atlas with Our RCCV.} To verify our proposed novel RCCV is a general 3D geometric feature representation and not limited to a specific framework, we additionally apply it to Atlas~\cite{atlas} to replace its simple back-projected feature. The results meet our expectations.
    }

\label{tab:atlas}
\end{table}

\textbf{Ray Compensation.} As analyzed in section~\ref{sec:rccv}, a single value in the standard cost volume is not sufficient to decode the geometry without its camera ray distribution as a reference. Our experiment in the second group verifies this assumption by showing a noticeable performance boost with the Ray Compensation module. As shown in Fig~\ref{fig:floating}, our ray compensation module solves the floating artifacts problem and generates more clear geometry.

\textbf{Contextual Compensation.} In the last group, we explore the necessity of our proposed contextual compensation mechanism. 3 sets of experiments are designed as follows: (1)We simply concatenate the keyframe 2D image feature to the Ray-compensated Cost Volume. All 3D mesh metrics are improved by this concatenation, confirming the effectiveness of the contextual feature. (2)A 2D convolutional layer is employed to better fuse the contextual feature with the cost volume, the improvements are relatively minor. (3)We use a group convolution layer to limit the perceptive field of the depth planes in the cost volume to be within itself. A more significant boost of the F1-score is observed, which shows the convolution layer is unable to preserve the structural order of the depth planes and manual regularization is necessary.

\label{sec:atlas}
\textbf{Downstream Module Agnostic.} Our proposed $RCCV$ is a general 3D geometric feature representation and is agnostic to the downstream fusion and prediction modules. The above experiments employ the VoRTX~\cite{vortx} for the downstream prediction, which is similar to other models like the NeuralRecon~\cite{neuralrecon} and TransformerFusion~\cite{transformerfusion}. To further verify the generalization ability of our $RCCV$, we apply it to the Atlas~\cite{atlas} framework to replace its simple back-projected feature. As shown in Table~\ref{tab:atlas}, our method improved all evaluation metrics by a large margin.

\section{Additional Discussions}
\subsection{Information Lost of Depth-based Methods}
The cost volume is widely used in depth-based reconstruction methods. Compared to their existing 3D-2D-3D pipelines, our end-to-end 3D volumetric reconstruction from the cost volumes have several fundamental advantages. In addition to the qualitative and quantitative evaluations in the main paper, here we analyze a typical case in the ScanNet2~\cite{scannet} dataset testing split.

In the supplementary material, we visualize the meshes, point clouds, a sample keyframe, and the matching confidence distribution of a sample pixel from the state-of-the-art depth-based method SimpleRecon~\cite{simplerecon}. The ground truth depth is 3.3 meters for the sample pixel, which is correctly reflected by its overall matching confidence distribution. However, SimpleRecon mistakenly predicts a depth of 2.92 meters due to a glitch in the cost distribution. Since the cost distribution information is discarded after the depth prediction, the downstream TSDF Fusion~\cite{kinectfusion} is unable to filter out this outlier and generates a floating surface artifact. In contrast, our end-to-end 3D volumetric framework preserves the cost volume information of all the keyframes and holistically reconstructs clear geometry.

\subsection{Use of Reference Frames}
The construction of our keyframe cost volumes requires reference image frames. While most of these reference images come from the keyframe pool, our method may utilize slightly more image frames than some volumetric baselines, depending on the chosen frame selection strategy. To determine if our improved performance is due to this additional information, we conducted two experiments. (1) As mentioned in Sec $4.3$ of the main paper, we apply our $RCCV$ to the Atlas~\cite{atlas} baseline. Both the baseline and our modified Atlas were using all available image frames, ensuring a fair comparison. (2) We evaluated our baseline VoRTX~\cite{vortx} using the same frames as our method and found that the F1-Score only improved from 0.703 to 0.705, indicating a negligible difference in performance that does not affect our conclusions.

\subsection{Computation Efficiency}
Constructing cost volumes requires additional computation time and memory compared to existing volumetric baselines. In our experiment, we find reducing the channel number of the cost volume from $7$ to $1$ and the number of depth planes from $64$ to $32$ does not noticeably affect reconstruction quality but will greatly reduce the computation overhead. The $RCCV$ of $R^{32 \times 1 \times 60 \times 80}$ only consumes $300KB$ of the memory and $5ms$ of the GPU time.

\subsection{RCCV fusion strategy:}
We adopted the VoRTX~\cite{vortx} (attention-based) and Atlas~\cite{atlas} (averaging) way of fusion strategy and both of them work great with our RCCV. We will clarify this in the future version.\\

\subsection{Limitations}
One major limitation of volumetric reconstruction methods like ours is the update of results is slower than depth-based methods, which could be alleviated by a proper fragmenting strategy.

\section{Conclusion}
In this paper, we identify fundamental limitations of the existing neural reconstruction methods and present an end-to-end 3D reconstruction framework named CVRecon with novel cost-volume-based 3D geometric feature representation $RCCV$. We significantly outperformed existing state-of-the-art methods and provide valuable insights into the development of 3D geometric feature learning schemes.

\clearpage
\clearpage

{\small
\bibliographystyle{ieee_fullname}
\bibliography{egbib}
}

\end{document}